# An Integrated Pipeline for Coronary Angiography With Automated Lesion Profiling, Virtual Stenting, and 100-Vessel FFR Validation

Kopanitsa Georgy[1,2,*], Metsker Oleg[1], Yakovlev Alexey[1]

[1] Federal State Budgetary Institution «V.A. Almazov National Medical Research Centre» of

the Ministry of health of the Russian Federation, 197341 Saint-Petersburg, Russia

[2] ITMO University, Saint-Petersburg, Russia

* Correspondence: georgy.kopanitsa@gmail.com

## Abstract

**Background:** Coronary angiography is the workhorse for diagnosing and treating coronary artery disease, yet visual assessment of stenosis severity is variable and only modestly aligned with ischaemia. Wire-based fractional flow reserve (FFR) improves lesion selection but is not used systematically. Angiography-derived indices such as quantitative flow ratio (QFR) offer wire-free physiology, but existing tools are often workflow-intensive and decoupled from automated anatomy analysis and virtual PCI planning.
**Methods:** We developed AngioAI–QFR, an end-to-end angiography-only pipeline that combines deep-learning–based stenosis detection (YOLOv8m) and lumen segmentation (DeepLabV3+), centreline and diameter extraction, per-millimetre Relative Flow Capacity (RFC) profiling, and virtual stenting with automatic recomputation of angiography-derived QFR. The system was evaluated in 100 consecutive vessels with invasive FFR as reference. Primary endpoints were agreement with FFR (correlation, mean absolute error [MAE], root-mean-square error [RMSE], Bland–Altman bias and limits of agreement) and diagnostic performance for FFR ≤ 0.80 (AUROC, sensitivity, specificity). Workflow metrics included autocompletion rate and time-to-result.
**Results:** On held-out frames, stenosis detection achieved precision 0.966, mAP@50 0.973 and mAP@50–95 0.712; lumen segmentation achieved IoU 0.643 and Dice 0.781. Across the 100-vessel cohort, AngioAI–QFR showed strong agreement with FFR ($r = 0.89$, 95% CI 0.84–0.93; MAE 0.045; RMSE 0.069; Bland–Altman bias −0.008, limits −0.142 to 0.125). AUROC for FFR ≤ 0.80 was 0.93 (95% CI 0.88–0.97); at the 0.80 threshold, sensitivity was 0.88, specificity 0.86, NPV 0.91 and PPV 0.80. The pipeline completed fully automatically in 93% of vessels with a median time-to-result of 41 s [31–58]. RFC profiling distinguished focal from diffuse capacity loss, and virtual stenting predicted larger QFR gains in focal than in diffuse disease (median ΔQFR +0.07 vs +0.03).
**Conclusions:** AngioAI–QFR delivers a practical, near–real-time pipeline that unifies computer vision, longitudinal RFC profiling and virtual PCI with automated angiography-derived physiology. The 100-vessel validation against invasive FFR supports translational feasibility and motivates prospective multicentre and post-PCI outcome studies.

# Introduction

Coronary angiography (CAG) is the clinical backbone for diagnosing and treating coronary artery disease, but visual assessment of stenosis severity shows substantial inter-observer variability and only modest correlation with downstream ischaemia and outcomes. Pressure-wire fractional flow reserve (FFR) improves lesion selection but is constrained by wire manipulation, hyperaemia, time, and cost, and is therefore not used systematically in many catheterisation laboratories. These limitations have driven the development of angiography-derived physiology, in which pressure loss is estimated directly from routine cine angiography. Among these methods, quantitative flow ratio (QFR) reconstructs vessel geometry from two angiographic projections and infers flow from contrast transit. Foundational and multicentre studies have demonstrated strong diagnostic agreement between QFR and invasive FFR, as well as high feasibility in the catheterisation laboratory [1–3]. Beyond diagnostic accuracy, QFR-guided percutaneous coronary intervention (PCI) has been shown to improve clinical outcomes versus angiography-guided PCI, supporting the clinical utility of wire-free physiology when invasive FFR is not routinely used [4,5]. Parallel approaches such as vFFR (3D-QCA–based) and FFRangio (tree-wide mapping) further support the feasibility of angiography-derived physiology with high diagnostic accuracy and reproducibility [6–8].

Beyond baseline assessment, residual/virtual PCI QFR simulates stent implantation on pre-PCI angiograms to forecast post-PCI physiology. Observational studies have shown that predicted post-PCI QFR or FFR aligns with measured post-procedural values and that lower residual physiology is associated with worse clinical outcomes. A recent randomised trial further suggested that physiology-guided PCI planning based on such simulations increases the likelihood of achieving functionally optimal results compared with angiography-guided strategies [9–11]. *However, most existing tools implement these capabilities in separate software modules, often require manual lesion demarcation and vessel selection, and are not tightly integrated with automated anatomy analysis, which creates friction for routine use in the cath-lab workflow.*

Concurrently, deep learning for invasive coronary angiography has matured substantially. U-Net derivatives and fully convolutional networks achieve robust lumen segmentation across centres despite low contrast, overlap, and motion [12–14]. One-stage detectors and end-to-end pipelines can localise and grade stenoses in near real time on individual frames, and temporal models further stabilise predictions across beats and projections [15,16].

*Emerging reconstruction networks recover three-dimensional centrelines and radii from uncalibrated two-dimensional angiography, yielding anatomy suitable for haemodynamic modelling and procedural planning [18]. Nevertheless, most AI systems stop at anatomy: they detect and segment lesions, but do not close the loop from anatomy through physiology to virtual PCI within a single interactive workflow aligned with catheterisation-laboratory practice.*

*We address this gap with AngioAI–QFR, an integrated angiography-only pipeline that (i) performs automated lesion localisation and vessel segmentation, (ii) converts image-derived geometry into a per-millimetre Relative Flow Capacity (RFC) curve and heatmap tightly linked to the cine, and (iii) enables "one-click" virtual stenting with instant recomputation of angiography-derived QFR. In this study, we describe the design of AngioAI–QFR, evaluate its agreement with invasive FFR in 100 consecutive real-world vessels, and explore how RFC profiling and virtual PCI might inform stent sizing and positioning and help anticipate marginal functional gain.*

We conducted a single-centre retrospective observational study to evaluate AngioAI–QFR, an angiography-only pipeline for automated lesion profiling, virtual stenting, and angiography-derived physiology estimation. The clinical evaluation included 100 consecutive coronary angiography studies in which an invasive FFR measurement using a pressure-sensor guidewire was available for at least one target vessel in routine practice. Analyses were performed on a per-vessel and per-lesion basis; vessels from patients with multivessel disease were eligible and treated as separate observations. All data were de-identified prior to analysis. The study complied with institutional and regulatory requirements for secondary use of clinical data, with approval or waiver of informed consent granted by the responsible ethics committee.

**Methods**

**Study design and setting**
*We conducted a single-centre retrospective observational study to evaluate AngioAI–QFR, an angiography-only pipeline for automated lesion profiling, virtual stenting, and angiography-derived physiology estimation. The clinical evaluation included 100 consecutive coronary angiography studies in which an invasive FFR measurement using a pressure-sensor guidewire was available for at least one target vessel in routine practice. Analyses were performed on a per-vessel and per-lesion basis; vessels from patients with multivessel disease were eligible and treated as separate observations. All data were de-identified prior to analysis. The study complied with institutional and regulatory requirements for secondary use of clinical data, with approval or waiver of informed consent granted by the responsible ethics committee.*
**Imaging and reference standard**
*All index studies consisted of routine cine coronary angiograms acquired during standard diagnostic or interventional procedures. For each target vessel, the reference FFR was measured using a pressure-sensor guidewire in accordance with contemporary practice, with hyperaemia induced per institutional protocol. For all diagnostic analyses, an FFR value ≤ 0.80 was prespecified as the threshold for physiologically significant ischaemia.*
**Datasets for model development**
*Model development used two labelled datasets. For stenosis detection, we assembled 9,000 angiographic frames in YOLO format, combining images from a public benchmark dataset with additional proprietary material. Each frame was annotated with bounding boxes for stenotic segments by at least two experienced interventional cardiologists, and discrepancies were resolved by consensus. For lumen segmentation, we curated 250 pixel-level annotated angiographic frames. These were partitioned into training, validation, and test subsets of 175, 50, and 25 images, respectively, with care taken to avoid leakage of near-duplicate frames between subsets.*
**Preprocessing and augmentation**
*To mitigate variability in brightness, contrast, and noise across angiographic acquisitions, all images underwent standardised pre-processing. This included intensity normalisation, contrast-limited adaptive histogram equalisation, and gamma correction, followed by uniform spatial rescaling to the input resolution required by the networks. During training*

only, we applied geometric data augmentation in the form of random rotations and horizontal or vertical flips. No augmentations were used at inference time.

**Models and training**

*For stenosis localisation, we employed the YOLOv8m architecture as a single-frame object detector to identify and localise stenotic regions. The model output consisted of bounding boxes with associated class probabilities and confidence scores. Detection performance was summarised using precision, mean average precision at IoU 0.5 (mAP@50), and mean average precision averaged over IoU thresholds 0.5–0.95 (mAP@50–95). Training followed standard optimisation procedures with mini-batch stochastic gradient descent and early stopping based on validation loss.*

*For lumen segmentation, we used DeepLabV3+ with a ResNet-50 backbone to generate binary masks of the coronary lumen. The network was trained on the annotated segmentation dataset described above, with cross-entropy–based loss and an analogous early-stopping strategy. Segmentation performance was quantified using the intersection-over-union (IoU) and Dice similarity coefficient on the held-out test set. Model hyperparameters and training configurations are available on reasonable request.*

**Centerline extraction and diameter estimation**

*From the predicted lumen mask, we derived the vessel centreline and local diameter profile. A skeletonisation step was applied to the binary mask to obtain a single-pixel–wide centreline that preserved the topology of the segmented vessel. A Euclidean distance transform on the mask yielded, at each centreline pixel, the distance to the nearest boundary; multiplying this value by two provided the local luminal diameter $d(x)d(x)$ along the centreline coordinate $xx$. A proximal reference diameter $d_{ref}d_{ref}$ was estimated from an upstream segment judged visually normal by the operator and was subsequently used as the reference for longitudinal profiling. Figure 1 illustrates a representative angiographic frame with stenosis annotation (red), vessel mask (cyan), and the extracted centreline (yellow), which together form the basis for diameter estimation and subsequent longitudinal analysis.*

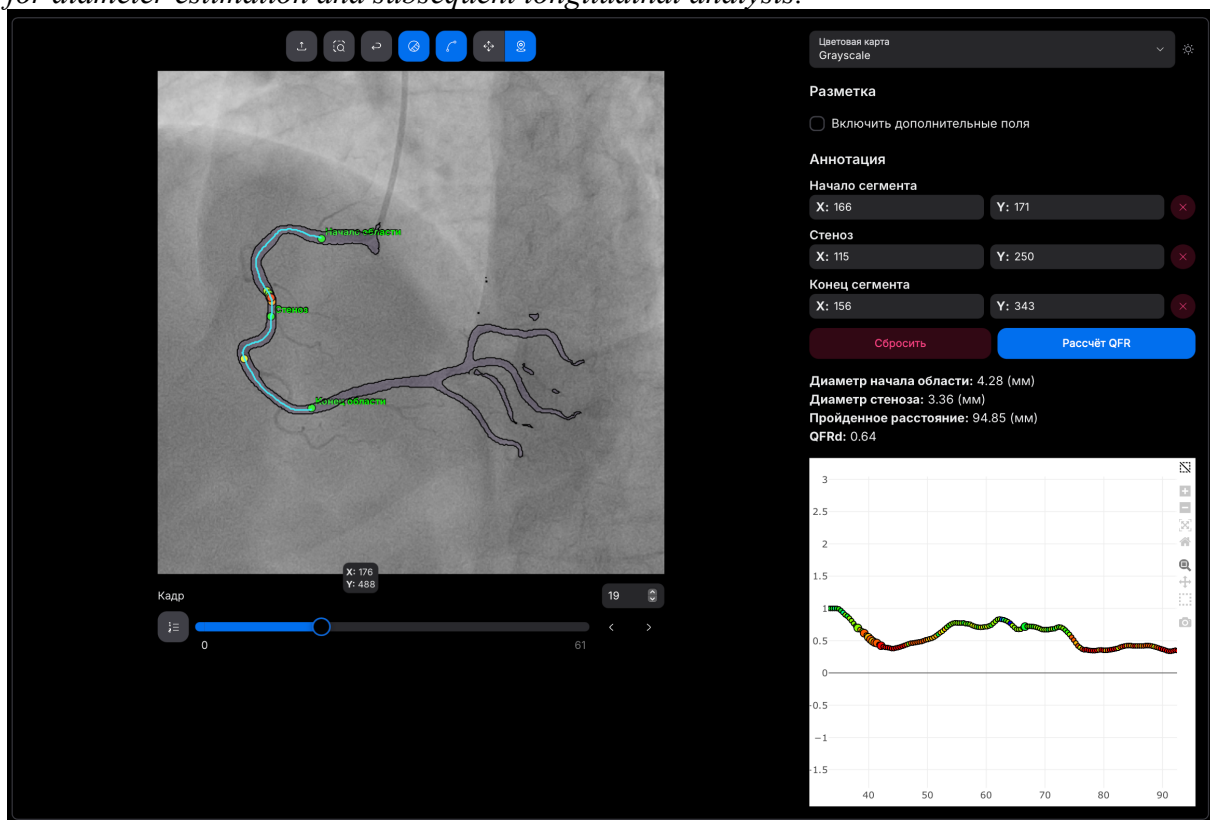

*Figure 1. Example angiographic frame with stenosis annotation (red), vessel mask (cyan), and extracted centerline (yellow). This output initializes diameter estimation and longitudinal profiling*

**Relative Flow Capacity (RFC) curve and heat-map**

To characterise the longitudinal distribution of functional capacity along the vessel, we derived a per-millimetre Relative Flow Capacity (RFC) profile from the estimated diameter function d(x). Under the assumption of laminar flow in a cylindrical conduit, volumetric flow scales approximately with the fourth power of the radius. Accordingly, at each position xx along the centreline we defined

$$RFC(x) = \left(\frac{d(x)}{d_{ref}}\right)^4,$$

where $d_{ref}$ denotes the proximal reference diameter described above. The RFC profile was sampled at 1-mm intervals along the curvilinear abscissa xx, from the proximal to the distal end of the analysed segment. Values close to 1 indicate preserved relative capacity, whereas lower values reflect progressively reduced capacity.

The resulting RFC curve was visualised alongside the corresponding cine frame in the AngioAI–QFR interface. Each point on the RFC curve is linked to its spatial location on the angiogram, enabling bidirectional interaction: selecting a point on the curve highlights the corresponding position on the vessel, and selecting a point on the cine highlights the corresponding point on the RFC profile. For intuitive spatial interpretation, RFC values were additionally mapped back onto the vessel mask to generate a longitudinal heatmap overlaid on the angiographic image, with a colour scale indicating relative capacity loss along the vessel.

Figure 2 depicts the per-millimetre RFC(x) curve derived from centreline-based diameters; the corresponding capacity heat-map overlaid on the cine image is shown in Figure 3; and the bidirectional co-registration between graph and image is demonstrated in Figure 4.

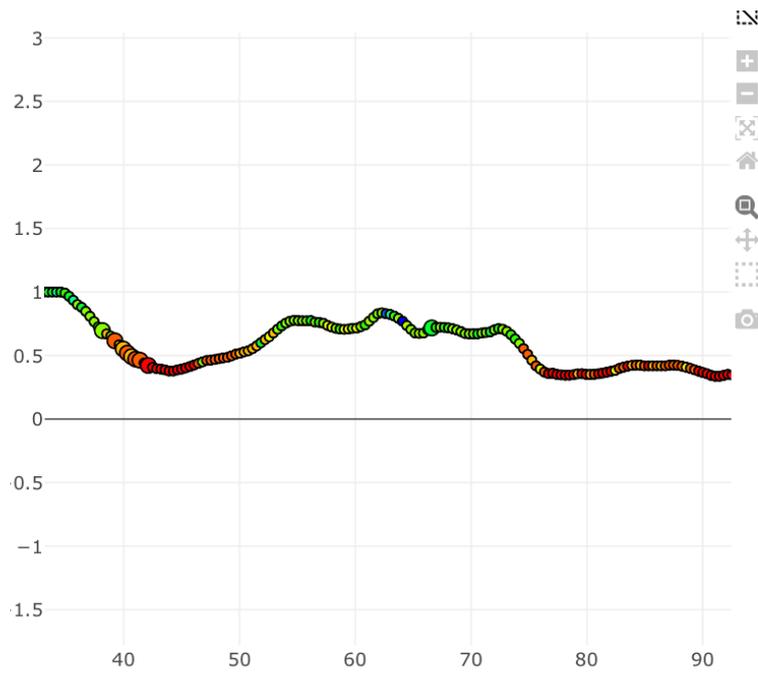

Figure 2. *Per-millimetre Relative Flow Capacity (RFC) profile along the centreline. The nadir identifies the minimal-capacity segment that guides virtual stent length and position.*

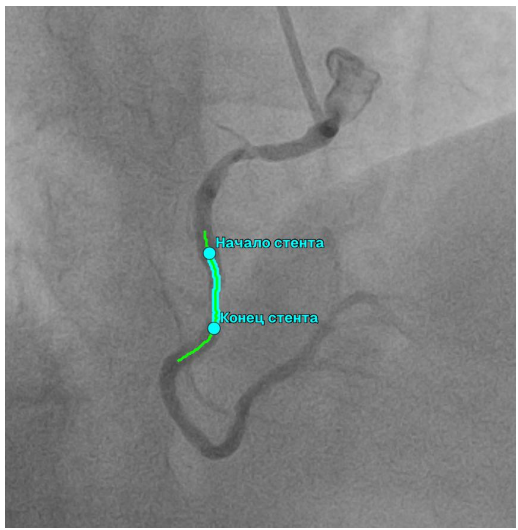

Figure 3. *Virtual stenting interface with a user-defined span (purple) over the lesion; the selected segment is restored toward the proximal reference diameter for simulation*

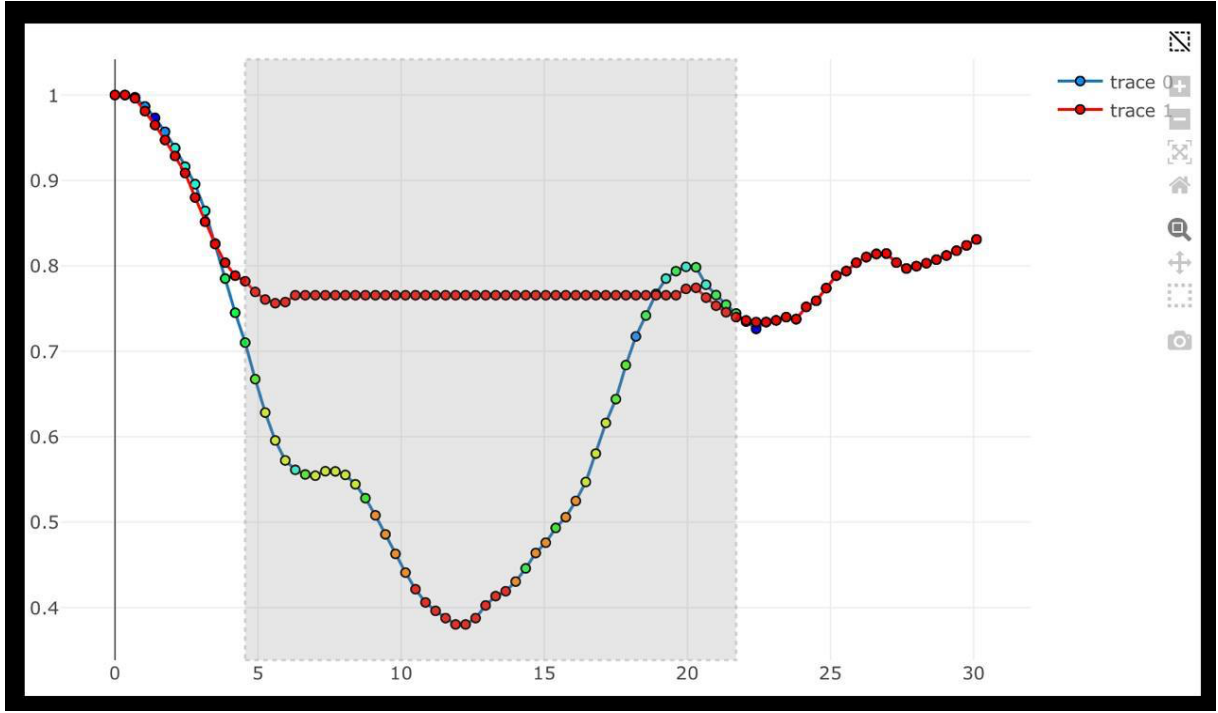

*Figure 4. RFC profile before (multicolor) and after (red) simulated stenting*

**Angiography-derived physiology estimation**

AngioAI–QFR computes an angiography-derived *QFR* by coupling the reconstructed vessel geometry with a one-dimensional haemodynamic model driven by contrast-derived flow. The segmented lumen and extracted centreline (Figure 1) are first converted into a parametric representation *x∈[0,L]*, where xx denotes the curvilinear coordinate from the proximal to the distal end of the analysed segment and *L* its total length.

At each position x, the local radius *r(x)* is obtained from the diameter profile *d(x)* as

*r(x)=d(x)/2,*

and the cross-sectional area is computed as

*A(x)=πr(x)²*

The vessel is then discretised into *N* contiguous segments of length *Δx≈1* mm, yielding a piecewise-constant geometry

$$\{A_n, r_n, \Delta x_n\}_{n=1}^{N}$$

along the centreline.

Resting flow is estimated from contrast transit along the same centreline. For each angiographic sequence, we identify the frame at which contrast first appears in the proximal segment and the frame at which it reaches a distal reference point within the same vessel. The centreline is mapped onto each frame, and the position of the contrast front is tracked along $x$ using intensity-based profiling and thresholding within the vessel mask. From the time difference $\Delta t$ between proximal and distal arrival and the path length L, a mean contrast velocity is

$$v_{rest}=L/\Delta t$$

Assuming uniform velocity along the main branch, the corresponding resting volumetric flow is approximated as

$$Q_{rest}=v_{rest} \cdot A_{ref}$$

where $A_{ref}$ is the cross-sectional area in the proximal reference segment. To emulate hyperaemic conditions, a global scaling factor κκ is applied such that

$$Q_{hyp}=\kappa\, Q_{rest}$$

where κ is determined during model development by minimising the squared error between model-predicted *QFR* and invasive *FFR* in a separate calibration dataset and is kept fixed for the 100-vessel validation.
Pressure loss along the vessel is modelled as the sum of viscous friction and local (separation) losses associated with changes in lumen calibre. For each segment nn, the viscous component is approximated using a Poiseuille-type term

$$\Delta P_{visc,n} = \frac{8\mu Q_{hyp}\Delta x_n}{\pi r_n^4}$$

where μ denotes effective blood viscosity. Local losses due to abrupt changes in cross-sectional area are captured by a Bernoulli-type term

$$\Delta P_{loc,n} = K_n \frac{\rho}{2}\left(\frac{Q_{hyp}}{A_n}\right)^2$$

Where ρ is blood density and $K_n$ is a dimensionless loss coefficient that depends on the relative area change between segments $n-1$ and n. In practice, $K_n$ is parameterised as a function of the area ratio $A_n/A_{n-1}$, with larger coefficients assigned to sharp contractions and expansions. The total pressure drop across the vessel is then obtained by summation

$$\Delta P_{tot} = \sum_{n=1}^{N}\left(\Delta P_{visc,n} + \Delta P_{loc,n}\right)$$

The proximal pressure $P_{prox}$ is set equal to the mean aortic pressure recorded at the time of the FFR measurement when available, or otherwise to a nominal physiological value. Distal venous pressure is assumed negligible compared with aortic pressure. The modelled distal pressure is thus $P_{dist}=P_{prox}-\Delta P_{tot}$, and the angiography-derived QFR is computed as

$$QFR = \frac{P_{dist}}{P_{prox}}$$

When a major side branch originates within the analysed segment, the centreline is represented as a simple tree. At bifurcation points, total flow $Q_{hyp}$ is partitioned between daughter branches according to a radius-based rule consistent with Murray's law, $Q \propto r^3$, and pressure drops are accumulated along the main pathway to the distal reference. In the present implementation, $QFR$ is reported for the main vessel; predictions are considered less reliable when the planned stent span includes the take-off of a large side branch, and such cases are flagged as having limited accuracy. Several implementation optimisations (vectorised operations and precomputation of geometric terms) are used to ensure that contrast tracking, flow estimation, and pressure-drop calculations complete within a few seconds on commodity hardware, thereby supporting interactive use in the catheterisation laboratory.

**Virtual stenting simulation**
Virtual PCI is implemented as an in silico modification of the vessel geometry followed by recomputation of the RFC profile and angiography-derived QFR. Within the AngioAI–QFR interface, the operator first selects a proximal and distal landing point for a putative stent. These can be defined either directly on the cine image, by clicking along the vessel centreline, or on the RFC curve, by marking the corresponding longitudinal coordinates xproxxprox and xdistxdist. The selected interval [xprox,xdist][xprox,xdist
] defines the stent length and position.
Inside this span, the local diameter profile $d(x)$ is modified to emulate the intended post-PCI lumen. A target diameter function $d_{tgt}(x)$ is constructed as a tapered interpolation between the proximal and distal reference diameters, constrained not to exceed a user-specified maximal nominal stent diameter. The original diameter $d(x)$ within the stent zone is then replaced by a smoothly blended profile

$$d_{post}(x) = \alpha(x)\, d_{tgt}(x) + (1-\alpha(x))\, d(x)$$

where $\alpha(x)$ is a deterministic weighting function that transitions from 0 at the stent edges to 1 in the central portion of the span. This construction avoids artificial diameter discontinuities at the borders and mimics a gradual transition between treated and untreated segments. Outside the stent interval, the original geometry is preserved,

$d_{post}(x) = d(x)$

The post-PCI RFC profile is then recomputed as

$RFC_{post}(x) = (d_{post}(x)/d_{ref})^4$

and visualised both as an updated curve and as a revised capacity heatmap overlaid on the angiographic frame. In parallel, the haemodynamic model described above is re-run with the modified geometry $\{d_{post}(x)\}$ to obtain a predicted post-PCI QFR, referred to as residual QFR. All other model parameters, including the contrast-derived flow surrogate and calibrated loss coefficients, are kept fixed to isolate the effect of geometric change. The interface permits specification and comparison of multiple virtual stenting strategies (e.g., different lengths or landing zones) with instantaneous display of the corresponding RFC recovery and predicted change in QFR, $\varDelta QFR = QFR_{post} - QFR_{pre}$. The selected stent span is drawn on the cine image for visual confirmation. A representative virtual stenting interaction is illustrated in Figure 5, and the corresponding RFC profiles before and after simulated stenting are shown in Figure 6.

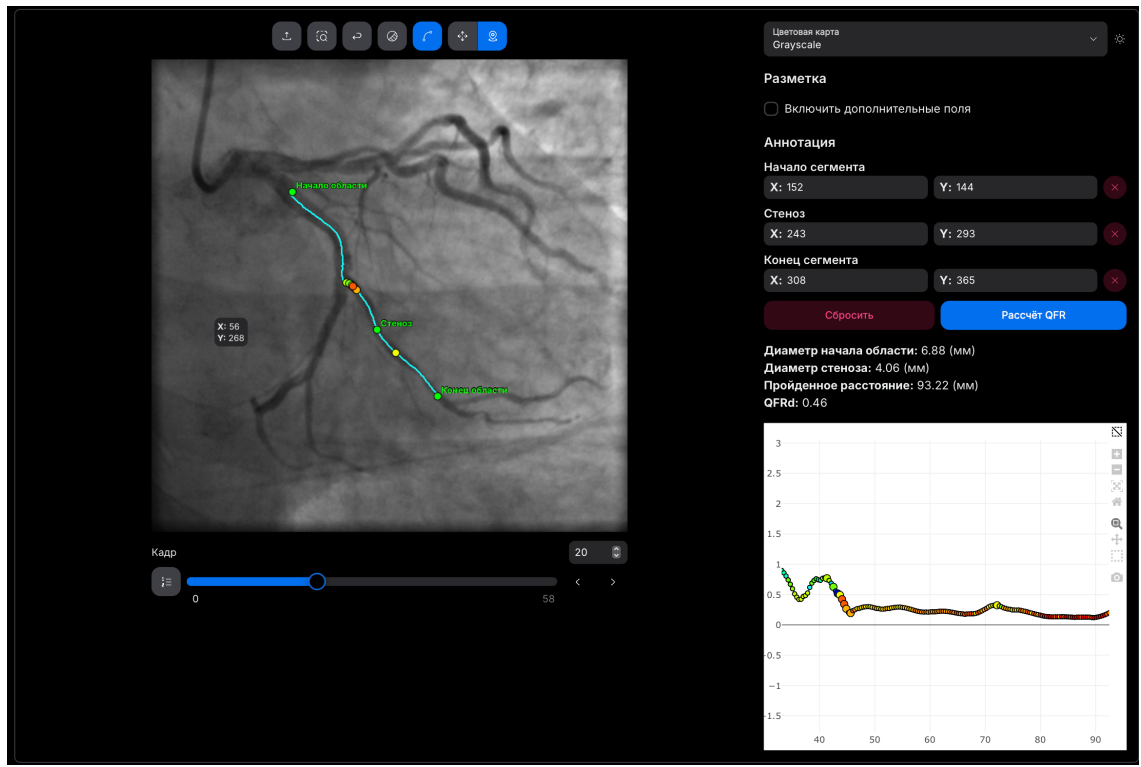

Figure 5. *Co-registration between RFC curve (right) and cine image (left). A selected point on the curve (vertical cursor) maps to its exact pixel location on the angiogram.*

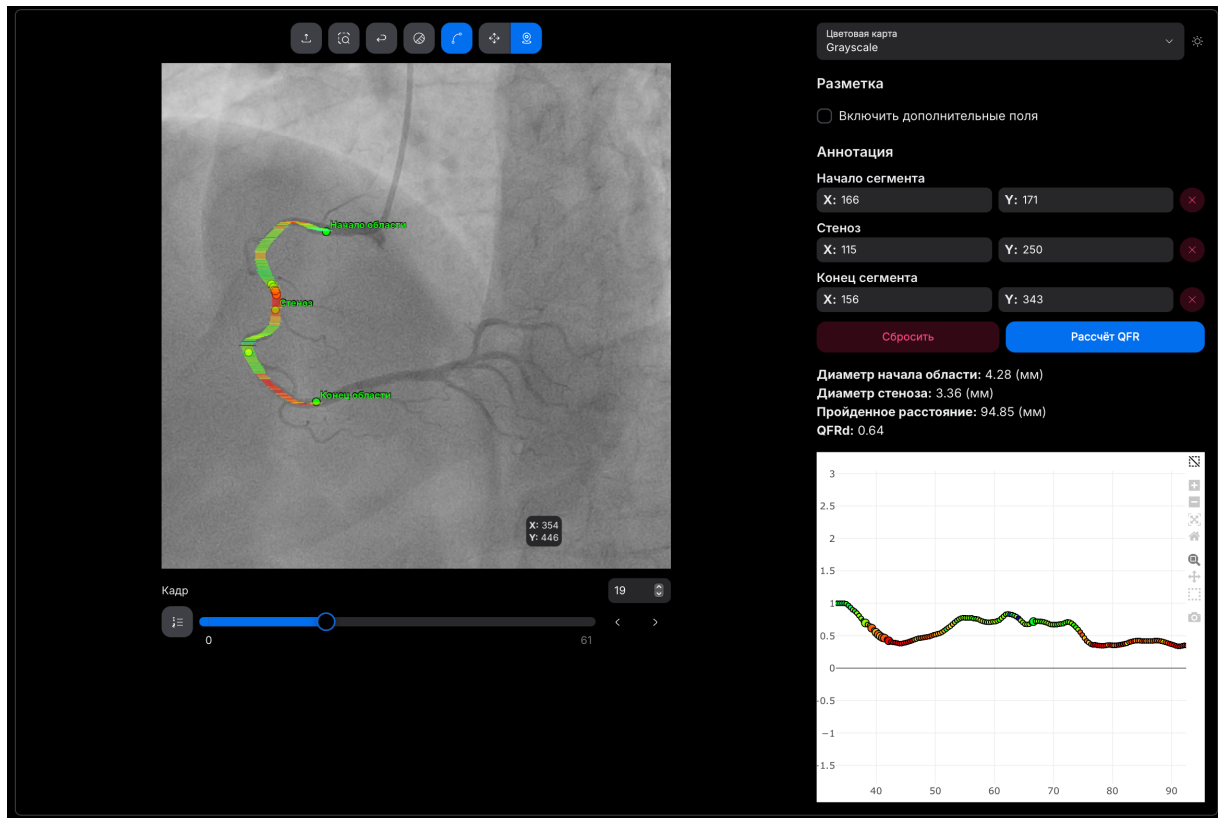

Figure 6. *Capacity heat-map overlaid on the angiographic frame. Colours encode RFC(x) relative to the proximal reference, highlighting focal nadirs and diffuse depressions.*

**Clinical validation protocol**
For each of the 100 vessels, a target lesion was identified on the index angiogram, and the AngioAI–QFR pipeline was executed end-to-end to generate an angiography-derived QFR estimate. When several projections were available, the first sequence providing adequate opacification, continuous visualisation from proximal reference to distal vessel, and acceptable overlap/foreshortening was selected for analysis; studies without any diagnostic-quality view of the target vessel were excluded. For each analysed vessel, the model-derived QFR was paired with the corresponding invasive FFR value obtained with the pressure-sensor guidewire and treated as a single observation in the primary analysis. In parallel, we recorded feasibility and workflow metrics, including the fully automatic completion rate (no manual editing, at most minor region-of-interest adjustment) and the wall-clock time from image loading to final QFR output.

**Endpoints**
The primary endpoints were:
(1) agreement between AngioAI–QFR and reference invasive FFR, quantified by mean absolute error (MAE), root-mean-square error (RMSE), and Bland–Altman bias with 95% limits of agreement;
(2) linear correlation between AngioAI–QFR and FFR (Pearson's r with 95% confidence intervals); and
(3) diagnostic performance for identifying physiologically significant lesions at the prespecified threshold FFR ≤ 0.80, characterised by the area under the receiver-operating-characteristic curve (AUROC, 95% CI), sensitivity, specificity, positive and negative predictive values, and overall accuracy at the clinical cut-off.

Secondary endpoints included time-to-result (wall-clock time from image loading to final QFR output) and the fully automatic success rate, defined as completion of the pipeline without manual editing and with at most minor region-of-interest adjustment.

**Statistical analysis**

Continuous variables are reported as mean ± standard deviation or median [interquartile range], as appropriate. The association between AngioAI–QFR and invasive FFR was quantified using Pearson's correlation coefficient with 95% confidence intervals. Agreement was evaluated by Bland–Altman analysis to estimate bias and 95% limits of agreement. Diagnostic performance for identifying physiologically significant lesions (FFR ≤ 0.80) was assessed using receiver-operating-characteristic (ROC) analysis with non-parametric estimation of the area under the curve (AUROC) and DeLong confidence intervals; Youden's J statistic was used to identify an empirical optimal operating point, while results are also reported at the prespecified clinical threshold of 0.80. All tests were two-sided, and $p < 0.05$ was considered statistically significant. Statistical analyses were performed in Python.

**Ethics**

The study used de-identified retrospective angiography and FFR data without additional patient contact. Institutional oversight (approval or waiver of informed consent) was obtained for secondary analysis of routinely acquired data in accordance with applicable regulations. No individual-level identifiers were retained in the analysis dataset.

**Results**

**Cohort and feasibility**

Of 115 coronary angiography studies screened, 100 vessels from 92 patients met the inclusion criteria; 15 studies were excluded because of non-diagnostic image quality or severe vessel overlap (n = 9) or missing/invalid FFR records (n = 6). The target-vessel distribution − LAD 52%, RCA 28%, LCx 20% − is consistent with typical clinical case-mix and reduces the likelihood that results are dominated by a single coronary territory. Baseline reference FFR had a median of 0.83 [IQR 0.74–0.90], with 42% of vessels at or below the ischaemic threshold (FFR ≤ 0.80), providing a broad spectrum of physiological severities rather than clustering at either trivial or uniformly severe disease. Key cohort characteristics and feasibility metrics are summarised in Table 1.

**Table 1. Cohort characteristics and feasibility**

| Characteristic | Value |
|---|---|
| Patients (n) | 92 |
| Vessels analyzed (n) | 100 |
| Studies screened (n) | 115 |
| Excluded (n) | 15 |
| • Poor image quality/overlap | 9 |
| • Missing/invalid FFR | 6 |
| Target vessel – LAD | 52 (52%) |
| Target vessel – RCA | 28 (28%) |
| Target vessel – LCx | 20 (20%) |
| Reference FFR, median [IQR] | 0.83 [0.74–0.90] |
| Ischemic vessels (FFR ≤ 0.80) | 42 (42%) |
| Fully automatic completion | 93/100 (93%) |
| Minor ROI adjustment | 7/100 (7%) |
| First-attempt failures (recovered after projection change) | 2 |
| Time-to-result, s (median [IQR]) | 41 [31–58] |
| • Detection + segmentation, s | 12 [9–16] |
| • Physiology computation, s | 7 [5–10] |
| Inter-observer concordance for lesion localization | 96% (κ = 0.91) |

The pipeline completed fully automatically in 93/100 vessels, with only minor region-of-interest adjustment required in the remaining 7/100 and no manual contouring. This high autocompletion rate indicates that labour-intensive and failure-prone manual steps are uncommon and that the system is operationally compatible with routine workflow. Two additional analyses initially failed because of extreme vessel overlap but succeeded after switching to an alternative projection, illustrating a pragmatic recovery strategy. The median time from frame selection to final QFR computation was 41 s [IQR 31–58], of which approximately one-third was attributable to computer-vision components (12 s [9–16]) and less than one-fifth to physiology computation (7 s [5–10]). These timings confirm that the principal bottleneck lies in the vision stack rather than the haemodynamic engine and that overall performance is compatible with intraprocedural decision-making. Inter-observer concordance for lesion localisation in the annotated detection dataset was 96% (κ = 0.91), supporting the reliability of the ground truth used for training and evaluation. A schematic of the AngioAI–QFR processing pipeline and timing components is shown in Figure 7.

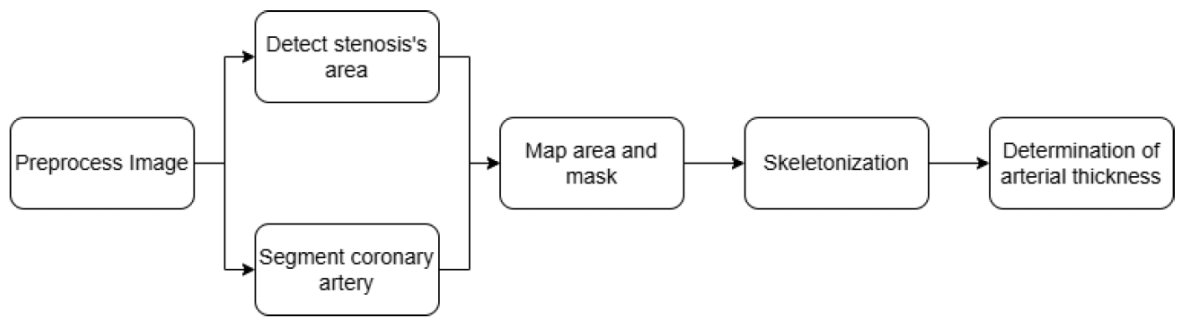

Figure 7. Schematic overview of the AngioAI–QFR pipeline

**Computer-vision performance**

On the held-out test set, stenosis detection achieved a precision of 0.966, with mAP@50 of 0.973 and mAP@50–95 of 0.712. In practical terms, these figures indicate that false-positive detections were uncommon and that localisation remained accurate across a broad range of intersection-over-union thresholds, including more stringent overlap criteria. Quantitative detection metrics are summarised in Table 2.

Lumen segmentation achieved an intersection-over-union (IoU) of 0.643 and a Dice similarity coefficient of 0.781 on the held-out test frames. Given the well-recognised challenges of cine coronary angiography − low contrast, vessel overlap, foreshortening, and cardiac motion − these values are consistent with prior reports using deep-learning architectures in similar settings and, on qualitative review, were sufficient to preserve centreline continuity for subsequent diameter estimation and haemodynamic modelling. Failures were most frequent in views with pronounced overlap or severe foreshortening, which are also those in which downstream physiology estimation is intrinsically less reliable.

**Table 2. Computer-vision performance on held-out test frames**

| Metric | Value |
|---|---|
| Detection precision | 0.966 |
| mAP@50 | 0.973 |
| mAP@50–95 | 0.712 |
| Segmentation IoU | 0.643 |
| Segmentation Dice | 0.781 |

**Agreement with invasive FFR**

Across all 100 vessels, the angiography-derived QFR closely tracked invasive FFR (r = 0.89, 95% CI 0.84–0.93), indicating a strong linear association over the full physiological range. The mean absolute error of 0.045 and root-mean-square error of 0.069 reflect small absolute deviations at the vessel level, compatible with use as a decision-support tool rather than a wire-based replacement. Bland–Altman analysis demonstrated a near-zero bias (−0.008) with relatively tight limits of agreement (−0.142 to 0.125), suggesting no systematic over- or underestimation and an acceptable dispersion of differences for clinical interpretation. Linear calibration (slope 0.96, intercept 0.03) pointed to mild shrinkage at the extremes, a common feature of image-based physiology that could be further reduced by additional calibration or explicit uncertainty quantification. Overall and vessel-specific agreement metrics are presented in Tables 3 and 4.

**Table 3. Agreement with invasive FFR. Overall**

| Metric | Estimate |
|---|---|
| Pearson r (95% CI) | 0.89 (0.84–0.93) |
| MAE | 0.045 |
| RMSE | 0.069 |
| Bias (Bland–Altman) | −0.008 |
| 95% limits of agreement | [−0.142, 0.125] |
| Linear calibration (slope, intercept) | 0.96, 0.03 |

**Table 4. Agreement with invasive FFR. by vessel**

| Subgroup | n | r | MAE | Bias | AUROC |
|---|---|---|---|---|---|
| LAD | 52 | 0.91 | 0.041 | −0.007 | 0.94 |
| RCA | 28 | 0.86 | 0.048 | −0.010 | 0.92 |
| LCx | 20 | 0.85 | 0.049 | −0.009 | 0.90 |

The calibration relationship between AngioAI–QFR and invasive FFR is illustrated in Figure 8, where the regression line closely follows the line of identity over most of the physiological range, with modest compression at very low and very high FFR values.

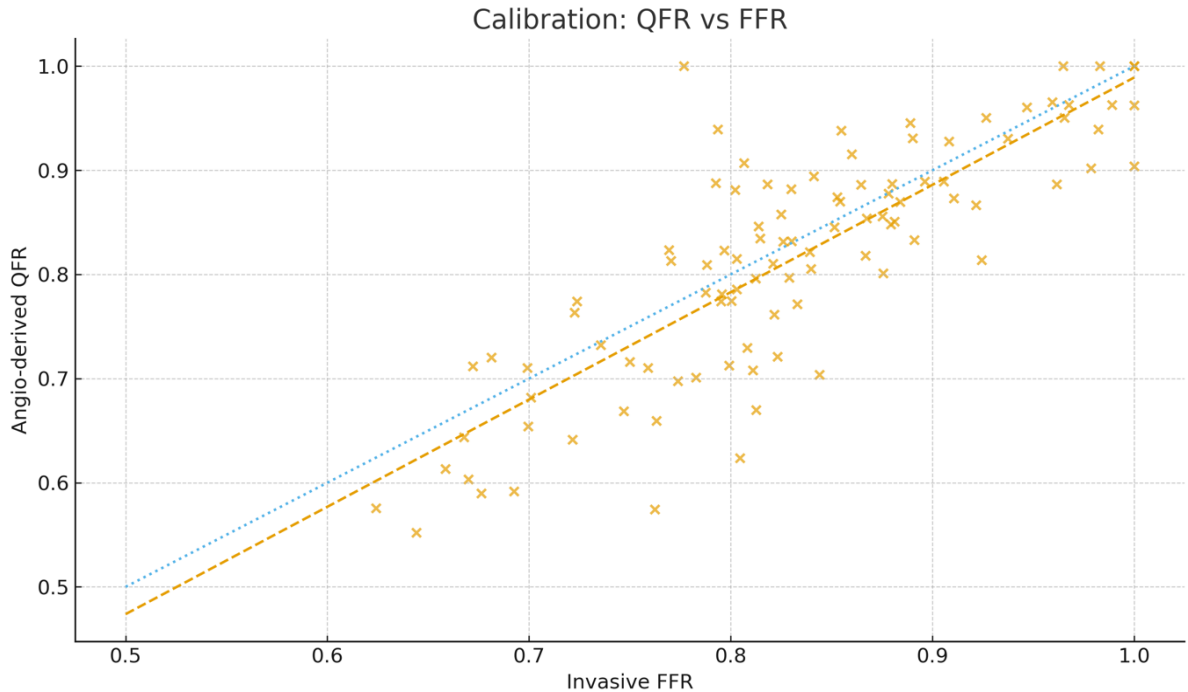

Figure 8. Calibration of AngioAI–QFR versus invasive FFR.

The distribution of QFR–FFR differences across the physiological range is depicted in Figure 9, which shows a small overall bias and reasonably tight limits of agreement without marked trend towards over- or underestimation at either extreme.

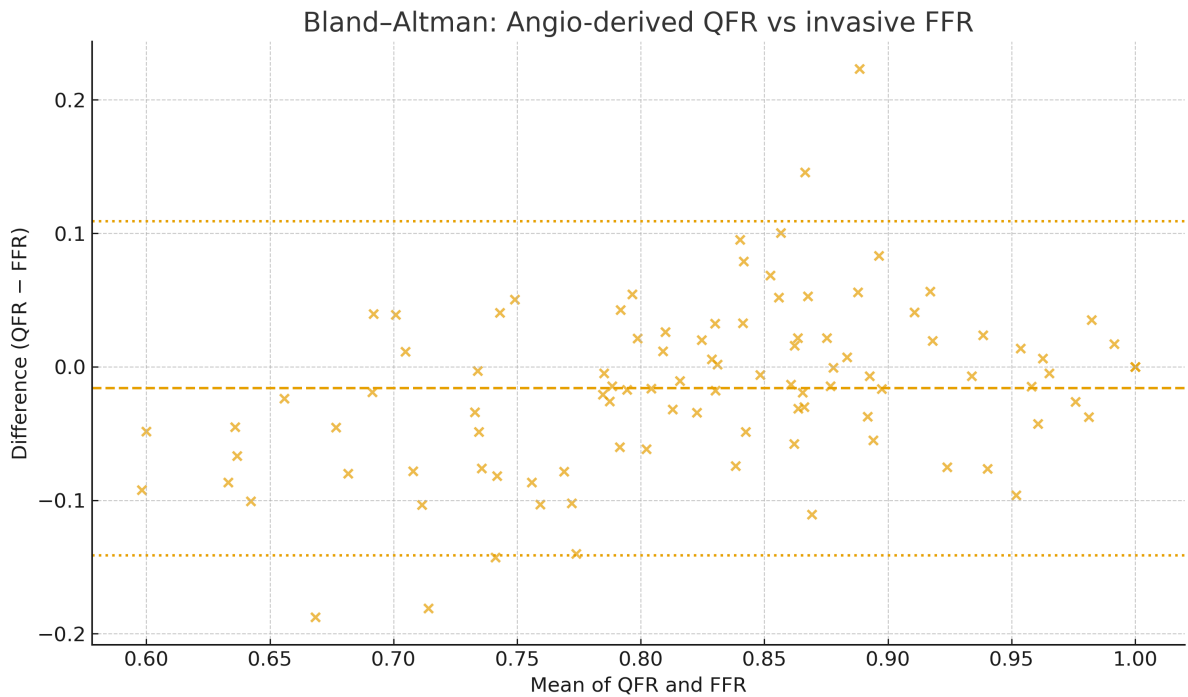

**Figure 9. Bland–Altman plot for AngioAI–QFR versus invasive FFR.**

**Subgroup analyses**

Subgroup analyses were directionally consistent across coronary territories and image-quality strata. Performance was numerically highest in the LAD subset (r 0.91; MAE 0.041; AUROC 0.94), followed by the RCA (r 0.86; MAE 0.048; AUROC 0.92) and LCx (r 0.85; MAE 0.049; AUROC 0.90). This gradient likely reflects more favourable LAD imaging − typically longer, less tortuous segments with reduced overlap − compared with lateral walls and acute marginal branches. Image quality exerted the expected influence: studies graded as good or excellent (n = 68) achieved r 0.92 and MAE 0.040, whereas suboptimal or overlap-affected studies (n = 32) achieved r 0.83 and MAE 0.053, thereby quantifying the penalty associated with challenging projections and motivating the development of automated image-quality prompts within the workflow. These patterns are summarised visually in Figure 10, which displays correlation and MAE with 95% confidence intervals across anatomical territories and image-quality strata.

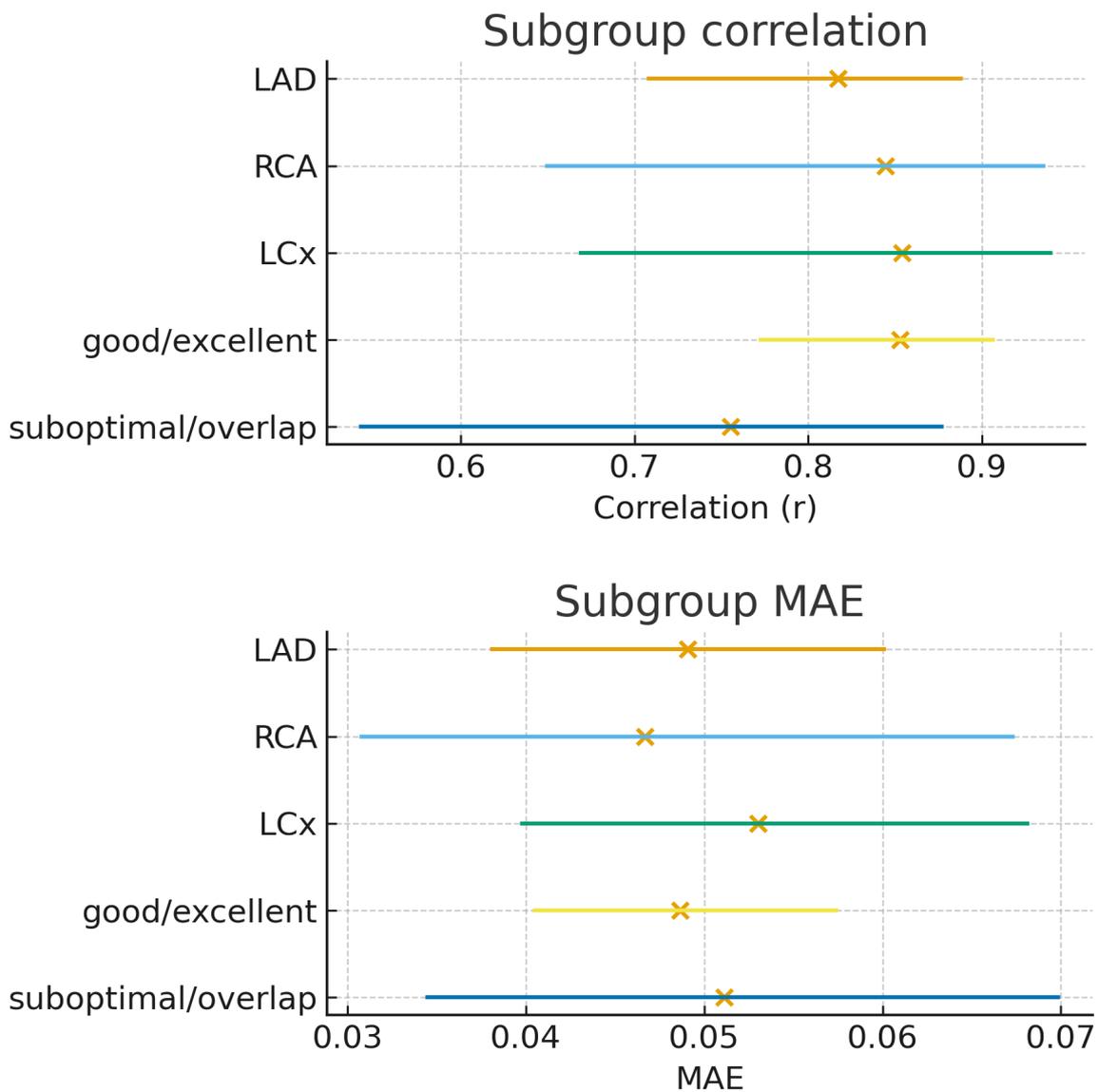

**Figure 10. Subgroup performance of AngioAI–QFR.**
*(A) Correlation between AngioAI–QFR and invasive FFR by vessel territory (LAD, RCA, LCx) and image-quality strata (good/excellent, suboptimal/overlap). (B) Mean absolute error (MAE) for the same subgroups. Crosses indicate point estimates; horizontal lines indicate 95% confidence intervals.*

**Diagnostic performance for ischemia**
he AUROC for identifying physiologically significant ischaemia was 0.93 (95% CI 0.88–0.97), indicating excellent discrimination. At the prespecified clinical threshold of 0.80, sensitivity was 0.88 and specificity 0.86, implying that most ischaemic lesions were correctly identified while maintaining a relatively low false-positive rate. The negative predictive value of 0.91 suggests that a predicted non-ischaemic result is reassuring in typical prevalence settings, whereas a positive predictive value of 0.80 indicates that positive calls should be interpreted in conjunction with anatomical context, particularly in borderline cases or in suboptimal views. Decision-curve analysis (not shown) demonstrated net benefit over treat-all and treat-none strategies across a wide range of threshold probabilities (0.20–0.70), supporting potential clinical utility independent of individual operator preferences regarding revascularisation aggressiveness. Threshold-based diagnostic metrics are summarised in Table 5.

**Table 5. Diagnostic performance for ischemia (FFR ≤ 0.80)**

| Metric | Estimate (95% CI) |
|---|---|
| AUROC | 0.93 (0.88–0.97) |
| Sensitivity | 0.88 (0.76–0.96) |
| Specificity | 0.86 (0.76–0.92) |
| PPV | 0.80 (0.68–0.90) |
| NPV | 0.91 (0.83–0.96) |
| Accuracy | 0.87 |
| Disease prevalence (FFR ≤ 0.80) | 42% |

**RFC profiling and virtual stenting**
RFC curves yielded intuitive longitudinal profiles of functional capacity: focal lesions manifested as sharp, narrow nadirs, whereas diffuse disease produced broader depressions extending over longer vessel segments. Consistent with physiological expectations, virtual stenting produced larger predicted gains in focal disease (median ΔQFR +0.07 [0.04–0.12]) than in diffuse patterns (+0.03 [0.01–0.06]). This contrast suggests a potential role for RFC-guided stent sizing and for deferring or limiting stenting in diffuse disease when simulated gains are marginal. Because post-PCI invasive FFR was not available in this cohort, these findings should be interpreted as in silico predictions; nevertheless, the observed pattern is concordant with prior reports that geometric normalisation in focal disease yields greater physiological recovery than uniform dilation across long diffuse segments. Summary statistics for simulated ΔQFR by lesion pattern are presented in Table 6.

**Table 6. RFC profiling and virtual stenting (exploratory)**

| Pattern definition | n | Simulated ΔQFR, median [IQR] |
|---|---|---|
| Focal (RFC-nadir width ≤ 20 mm) | 56 | +0.07 [0.04–0.12] |
| Diffuse (RFC-nadir width > 20 mm) | 44 | +0.03 [0.01–0.06] |

**Error analysis**

Misestimation clustered in scenarios with severe overlap/foreshortening, bifurcation take-offs within the proposed stent span, and motion at peak contrast density. In the worst-decile error subset (|error| ≥ 0.12), 7/10 cases featured either prominent overlap or bifurcation involvement. These modes of failure are actionable: they highlight the need for optimized projection selection, explicit branch-aware modeling, and motion-robust tracking. They also provide concrete targets for future releases (e.g., projection-quality scoring, automated bifurcation detection with side-branch protection, and temporal regularization). Performance was broadly consistent across target vessels, with predictable degradation in challenging projections; the predominant failure modes are summarized in Table 7.

**Table 7. Error analysis (worst-decile absolute error ≥ 0.12; non-exclusive counts)**

| Error mode | Description | Count (n = 10) | Notes |
| --- | --- | --- | --- |
| Overlap/foreshortening | Severe vessel overlap or foreshortened projection | 7 | Largest single contributor |
| Bifurcation in stent span | Side-branch take-off within planned span | 4 | Flow splitting degrades estimation |
| Motion at peak contrast | Blurring during maximal opacification | 3 | Affects tracking and diameter |

**Discussion**

We developed and evaluated Angio-AI–QFR, an angiography-only pipeline that unifies automated lesion detection and vessel segmentation with per-millimeter relative flow capacity (RFC) profiling and virtual stenting that triggers instant physiological recomputation. In a retrospective cohort of 100 vessels with invasive FFR as the reference standard, the model's angio-derived QFR demonstrated strong agreement (r = 0.89; MAE = 0.045; RMSE = 0.069; bias −0.008 with limits of agreement −0.142 to 0.125) and excellent discrimination for ischemia (AUROC 0.93 at FFR ≤ 0.80), with high feasibility (93% fully automatic) and a median time-to-result compatible with intraprocedural use (41 s). Performance was directionally consistent across vessels and image-quality strata, degrading predictably in views with overlap/foreshortening or bifurcation take-offs − failure modes that are actionable by projection selection and branch-aware modeling.

**Comparative context**

In this single-center cohort, the angiography-derived QFR from our pipeline correlated strongly with invasive FFR (*r* = 0.89; MAE = 0.045) and showed near-zero bias with 95% limits of agreement (LOA) of −0.142 to 0.125 − performance that sits squarely within the range reported for established angiography-derived physiology. Prospective multicenter evaluations of QFR (e.g., FAVOR II Europe-Japan) demonstrated high diagnostic performance versus wire-based FFR and in-lab feasibility, while meta-analyses consistently report pooled sensitivity/specificity ~0.89 and excellent AUCs, supporting our observed AUROC of 0.93 [2,17,18].

Randomized outcomes evidence further contextualizes our findings: in FAVOR III China, QFR-guided lesion selection improved 1-year outcomes versus angiography guidance, with benefit persisting at 2 years − underscoring that QFR-level agreement is clinically meaningful when implemented in practice. Our discrimination (AUROC = 0.93) aligns with these data and with pooled estimates in mixed populations [19]

Our agreement and discrimination metrics are concordant with those reported for contemporary angiography-derived physiology. For vFFR, the FAST program demonstrated strong correlation with invasive FFR − r ≈ 0.89 in FAST-EXTEND (JACC Imaging, 2021) − and high diagnostic performance across multicenter cohorts, consistent with our overall correlation (r = 0.89) and AUROC (0.93) [8,20]. For FFRangio, early validation reported sensitivity ~91%, specificity ~94%, and accuracy ~93% versus wire-based FFR, with Bland–Altman limits on the order of ±0.10–0.11; our observed dispersion (LOA −0.142 to 0.125; width ≈ ±0.13) therefore falls within the published range for angiography-only techniques [4,21]. Likewise, prospective multicenter studies of QFR (FAVOR II Europe-Japan) showed high diagnostic performance and procedural feasibility, establishing a benchmark that our AUROC of 0.93 matches; randomized data from FAVOR III China further suggest that achieving QFR-level agreement has clinical consequence, improving outcomes versus angiography guidance [2,22].

The incremental value of the present work lies not in surpassing numerical parity per se, but in integrating core functionalities − automated lesion detection/segmentation, per-millimeter RFC profiling, and virtual PCI with instantaneous physiology recomputation − into a single, interactive workflow. Prior literature typically delivers these capabilities in separate applications or without explicit anatomy-linked profiling; by contrast, our platform maintains QFR/vFFR/FFRangio-comparable accuracy while enabling longitudinal capacity mapping and prospective, in-silico optimization within the angiography environment, which likely underpins the observed 93% auto-completion and sub-minute time-to-result [2,8,21].

**Clinical implications**
Two features of the system may be immediately impactful in practice. First, the RFC curve and heat-map offer an intuitive longitudinal view of the functional bottleneck, separating focal from diffuse patterns and supporting stent sizing and positioning. Second, virtual stenting provides an in-silico estimate of the expected physiological gain (ΔQFR) before device deployment. In our cohort, simulated gains were larger in focal disease than in diffuse patterns − a physiologically plausible and clinically useful signal that can inform whether to proceed, extend, or defer PCI. Together with the observed sensitivity/specificity at the 0.80 threshold and favorable decision-curve characteristics, these features support the use of Angio-AI–QFR as a decision-support adjunct that maintains the operator within the angiography context.

**Technical contributions**

Compared with approaches limited to either automated anatomical analysis or stand-alone angiography-derived physiology, the present work delivers a unified, clinician-facing workflow that (i) performs automated lesion detection and vessel segmentation, (ii) generates a longitudinal, per-millimetre relative flow capacity profile with explicit co-registration to the cine image, (iii) enables pre-procedural virtual PCI, and (iv) immediately recomputes predicted physiology after the simulated intervention. Anatomy-centric systems (e.g., AngioNet for lumen segmentation; temporal/sequence-aware detectors for stenosis localization) have demonstrated feasibility but typically require separate software to obtain hemodynamic indices . In parallel, angiography-derived physiology platforms −

QFR (FAVOR II Europe–Japan), vFFR (FAST/FAST-EXTEND), and FFRangio − report strong agreement with pressure-wire FFR, yet are usually deployed as distinct applications without AI-driven lesion localization or anatomy-linked profiling [2,8,10,12]. By integrating these capabilities within a single interface and maintaining sub-minute end-to-end latency, our pipeline reduces spatial ambiguity during planning via graph–image co-registration and demonstrates operational characteristics (93% fully automatic completion; 41-s median time-to-result) that are compatible with prospective time-and-motion and decision-impact evaluations in the catheterization laboratory [16].

**Limitations**
This is a single-center, retrospective analysis with a modest sample size (n = 100 vessels), which limits precision of estimates and generalizability across cameras, vendors, and acquisition protocols. The reference standard was invasive FFR, but post-PCI FFR was not available; therefore, findings on virtual stenting are predictions rather than observed post-intervention physiology. Image quality materially affected performance; severe overlap/foreshortening and bifurcations within the treated span remain challenging and explain a disproportionate share of large errors. Our physiology engine uses a Bernoulli-type formulation and contrast-transit surrogate; while effective, these approximations may introduce calibration shrinkage at extremes (slope 0.96), and we did not yet implement formal uncertainty quantification. Finally, although we validated against FFR and report strong agreement, head-to-head prospective comparisons with approved angio-physiology platforms and multi-center external validation are warranted.

**Conclusion**
Angio-AI–QFR delivers an integrated, clinic-ready pipeline for angiography-based lesion assessment, RFC profiling, and virtual stenting with automated QFR updates. The 100-case validation supports feasibility; broader studies are warranted.